# A Unified Analytical Framework for Trustable Machine Learning and Automation Running with Blockchain

Tao Wang

*SAS Institute Inc.*

Cary, USA

t.wang@sas.com

*Abstract*—Traditional machine learning algorithms use data from databases that are mutable, and therefore the data cannot be fully trusted. Also, the machine learning process is difficult to automate. This paper proposes building a trustable machine learning system by using blockchain technology, which can store data in a permanent and immutable way. In addition, smart contracts are used to automate the machine learning process. This paper makes three contributions. First, it establishes a link between machine learning technology and blockchain technology. Previously, machine learning and blockchain have been considered two independent technologies without an obvious link. Second, it proposes a unified analytical framework for trustable machine learning by using blockchain technology. This unified framework solves both the trustability and automation issues in machine learning. Third, it enables a computer to translate core machine learning implementation from a single thread on a single machine to multiple threads on multiple machines running with blockchain by using a unified approach. The paper uses association rule mining as an example to demonstrate how trustable machine learning can be implemented with blockchain, and it shows how this approach can be used to analyze opioid prescriptions to help combat the opioid crisis.

## I. Introduction

Most machine learning algorithms have two common problems: trustability and lack of automation. First, it can be difficult to trust the results from a machine learning algorithm, because machine learning algorithms use data from databases that are mutable. System administrators and hackers can change the data source, and this will eventually change the results, with or without notification. From another perspective, it could be hard to understand why some machine learning algorithms are modeling the behavior of a system; but this is beyond the scope of the paper. Second, it is difficult to automate the machine learning process. Currently, the machine learning process is mostly controlled and monitored by human beings. Sometimes this process might begin or end at suboptimal times because of human involvement. This paper proposes building a unified analytical framework for trustable machine learning and automation running with blockchain [1] that can solve both problems.

A blockchain [1] is a continuously growing, single-linked list of immutable blocks that are often secured using cryptography. Blockchain was invented by Satoshi Nakamoto [1] in 2008 as a public financial transaction ledger for use in the cryptocurrency Bitcoin. Blockchain solved the Byzantine Generals Problem [3] by using a trustable system without going through a trusted financial institution. The double-spending problem [4, 5] was also solved in a purely peer-to-peer decentralized network without any financial institution involved. The network timestamps transactions by hashing them using SHA-256 [23] into an ongoing blockchain of hash-based proof-of-work (PoW), forming a record that cannot be changed without redoing the PoW (a.k.a. mining), which involves a substantial amount of computing power. The longest chain with the highest combined difficulties serves not only as proof of the sequence of transactions witnessed but also as proof that it came from the largest pool of computing power. With more and more computing power added to the blockchain every day, it is increasingly difficult to hack the blockchain system unless the hacker overpowers the rest of the world which is almost impossible in practice. So, people believe the data that are saved in blockchain are immutable and therefore trustable. This paper proposes that machine learning algorithms should use the immutable data provided by blockchain to solve the trustability problem.

While Bitcoin is widely considered to be blockchain 1.0, Ethereum [2] is considered to be blockchain 2.0, because it used blockchain not only as the foundation for cryptocurrency but also for DApps (Decentralized Applications) and DAOs (Decentralized Autonomous Organizations). The Ethereum network provided a blockchain with a built-in, fully fledged, Turing-complete [6] programming language that can be used to create so-called "smart contracts". Smart contracts are essentially some automated processes that can be used to encode arbitrary state transition functions, allowing users to create and run complicated systems (such as Facebook and Twitter, theoretically) on top of the Ethereum blockchain. The Ethereum blockchain opened a door to the largest development effort witnessed so far in the world of blockchain. There are many blockchain-based DApps in the making with ambitious goals to disrupt existing businesses. Therefore, some companies are looking at adopting blockchain with greater urgency. For instance, even some "traditional" companies like Kodak [9] are tapping into blockchain technology. This paper proposes to use the smart contract as the automation engine to solve the automation problem.

This paper makes three contributions. First, it establishes a link between machine learning technology and blockchain technology. Previously, machine learning and blockchain have been considered two independent technologies without an obvious link. Second, it proposes a unified analytical framework for trustable machine learning with blockchain technology. The proposed unified analytical framework solves both the trustability and automation issues in machine learning. Third, it makes it possible for a computer to translate (automatically or semi-automatically) core machine learning implementation from a single thread on a single machine to multiple threads on multiple machines running with blockchain by using a unified approach. As an example, Association Rule Mining (ARM) [11] is used to demonstrate how trustable machine learning can be done with blockchain, and it shows how this technology can be used to analyze opioid prescriptions to help combat the opioid crisis.

The rest of this paper is organized as follows. Section II reviews the background and prior art. Section III presents the proposed unified analytical framework for trustable machine learning and automation running with blockchain. Experiment and results are presented in Section IV. Section V concludes the paper and points out some future research directions.

## II. BACKGROUND AND LITERATURE REVIEW

Combining blockchain technology with machine learning technology is a very new idea. This makes it difficult to find closely related papers in the literature. In fact, only one such manuscript has been identified. In that manuscript [22], the authors proposed a so-called DanKu (Daniel + Kurtulmus) protocol to enable users to solicit machine learning models for a reward on the Ethereum [2] blockchain. The DanKu protocol has three phases. In the first phase, a reward giver submits a data set, an evaluation function, and a reward amount to the Ethereum contract, and solicits the best machine learning model for this data set with the given evaluation function. In the second phase, machine learning model providers download the data set and work independently to train a machine learning model that can represent these data. When a provider succeeds in training a model, he or she submits the model to the Ethereum blockchain. In the third phase, the Ethereum blockchain evaluates the submitted models by using the evaluation function and selects a winner. Unfortunately, the DanKu protocol works only on the Ethereum blockchain. Currently, two types of blockchains are widely used: public blockchains, such as Bitcoin [1] and Ethereum [2]; and permissioned blockchains (mostly for industry consortia), such as R3 Corda [7], Chain [8], BigChainDB [9], and Hyperledger [10]. It is unclear how the DanKu protocol would work on permissioned blockchains. Other issues raised by this manuscript [22] can be found in its section 6 ("Miscellanea and Concerns").

## III. A UNIFIED ANALYTICAL FRAMEWORK FOR TRUSTABLE MACHINE LEARNING AND AUTOMATION

There are many ways to classify machine learning algorithms. For example, machine learning algorithms can be grouped into four categories: supervised learning, unsupervised learning, semi-supervised learning, and others. Supervised learning requires the target (dependent) variable $Y$ to be labeled in the training data so that a model can be built to predict the label of unseen data, and it discards training observations that have unlabeled targets. Unsupervised learning does not require the target (dependent) variable $Y$ to be labeled in the training data, and its goal is not to predict the label but to infer a function to describe a hidden structure from unlabeled data. Semi-supervised learning requires only the target (dependent) variable $Y$ to be labeled in a small part of the training data; it is often used as a preprocessing step for supervised learning when labeling all the training data is impossible or too expensive. And there are many other machine learning algorithms, such as reinforcement learning, association rule mining [11], adversarial learning, and so on. While machine learning algorithms vary a lot, many of them have the following seven steps (assuming the data are already prepared):

1) Model Initialization. This is often the first step in many machine learning algorithms. In this step, the algorithm initializes the model by checking required licenses and system resources (such as GPU) and by allocating memory for the data and for necessary structures.

2) Model Training. In this step, the algorithm trains a machine learning model by using the training data set. This is usually one of the most important steps. The basic premise is that the trained model should be a good representation of the training data set.

3) Model Validation. This step can be optional, but it is crucial to many machine learning algorithms to prevent over-fitting. Over-fitting happens when the model memories the training data but also memories the noise. In this step, the algorithm validates the trained model on the validation (holdout) data set to prevent over-fitting so that the trained model can be applied to new or unseen data sets. Note that model validation can occur during model training, so the model knows when to stop. It is listed as a separate step here because model validation is less common in big data applications, mainly because of its repetitive nature and longer run time.

4) Model Scoring (Testing). This step is optional. In some unsupervised learning algorithms, there is no model scoring. For supervised learning, model scoring is very important and often required. For prediction and regression tasks, the basic idea is to use the trained model to score new or unseen data. Note that model scoring and testing can have different meanings in different contexts, but the terms are used interchangeably in the literature.

5) Model Evaluation (Assessment). This step is optional. The basic idea is to test how good the trained model is according to certain metrics. A machine learning algorithm can be compared with another one through model evaluation. Note that model evaluation and persistence can have different meanings in different contexts, but the terms are used interchangeably in the literature.

6) Model Serialization (Persistence). This step is optional. To ensure that a trained model can be used to score new or unseen data, it needs to be serialized and saved for future use. Note that model serialization and persistence can have different meanings in different contexts, but the terms are used interchangeably in the literature.

7) Model Clean-up. This is often the last step in many machine learning algorithms. In this step, the algorithm frees

the allocated memory and releases obtained system resources (such as GPU).

This paper uses these seven steps to demonstrate the unified analytical framework for trustable machine learning and automation running with blockchain.

As stated earlier, blockchain and machine learning are two completely different technologies without an obvious link. In this paragraph, a non-obvious link between the machine learning technology and the blockchain technology will be established. Although they are different, they have complementary characteristics. Machine learning builds probabilistic and ever-changing models based on the past or the current reality to predict the future. As we all know, machine learning needs deterministic and real data in order to build good models. Unfortunately, current data storage systems are all mutable, so the data can be modified by hackers or unauthorized parties or simply by mistake. Blockchain provides deterministic and permanent data that record reality and cannot be modified. As a result, machine learning algorithms can completely trust the data coming from blockchains and build good models. Therefore, blockchain holds the key to trustable machine learning.

Again, there are two major types of blockchains: public blockchains and permissioned blockchains. While people are familiar with public blockchains, many of them are not sure about the need for permissioned blockchains. A detailed discussion is beyond the scope of this paper, but a simple example can demonstrate that permissioned blockchains are in fact needed. For example, if a user (Alice) wants to sell a book to another user (Bob) with 20% discount and does not want to tell this discount to the public, then Alice should use permissioned blockchain to hide this discount information from the public. A machine learning algorithm can run on a blockchain or off a blockchain. When running off a blockchain, the algorithm needs to obtain data from the blockchain either at rest (in an aggregated way) or on the fly (in a streaming way). This paper covers both types of blockchain and different machine learning algorithm running modes, on or off a blockchain. More details are given later in this section. The concept is represented in Fig. 1.

As mentioned previously, a smart contract is an automated process that can be used to encode arbitrary state transition functions, allowing users to create and run complicated algorithms on a blockchain. One of the advantages of running machine learning algorithms on a blockchain is that there is no data movement. However, one disadvantage is that when thousands of machine learning algorithms are running on the same blockchain, they all consume CPU cycles, and the run time might become too long. Therefore, machine learning algorithms should sometimes be running off the blockchain. If the machine learning algorithm is running off the blockchain, the algorithm can either obtain all the data via a server layer at rest in an aggregated way or obtain the data one piece at a time on the fly via a streaming layer. In addition to using CPU, using GPU in machine learning is now the norm. Therefore, we assume GPU will be used when there is a fit and do not include either CPU or GPU in Fig. 1.

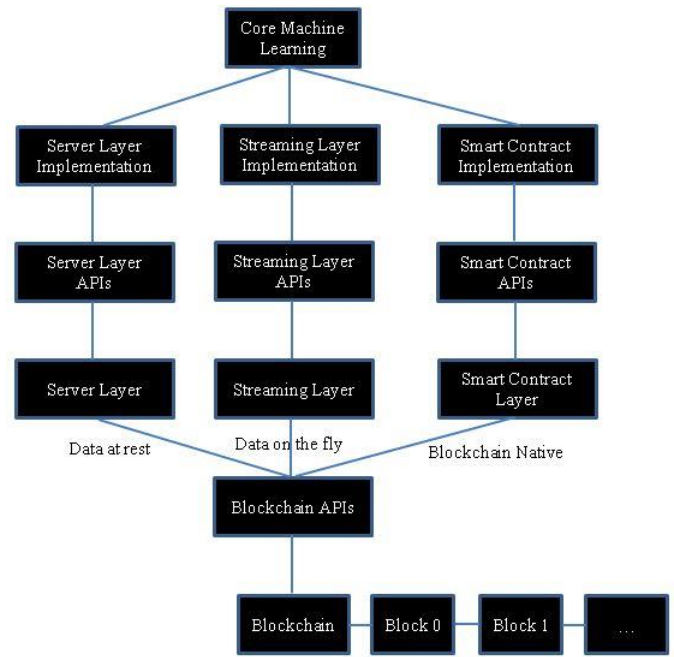

Fig. 1. The proposed analytical framework for trustable machine learning.

Note that this paper does not distinguish between computing nodes and consensus [1–2] nodes. There are good reasons to keep the computing process and consensus process on the same physical machine, and good reasons to separate them. This paper is generic, and therefore this discussion is beyond its scope. Note also that this paper is agnostic to the consensus algorithms [1–2], including proof-of-work algorithms, proof-of-stake algorithms, and others.

This framework is simple, but it covers almost all common blockchain variations, machine learning running modes, and their combinations. Therefore, it makes trustable machine learning and automation a reality. Currently, many machine learning algorithms are implemented in such a way that they can be running on top of the Server Layer (its concept will be explained later) to handle data at rest. They will have to be modified so that they can be running on top of the Smart Contract Layer and the Streaming Layer (their concepts will be explained later). Therefore, three versions of implementation (server implementation, streaming implementation and smart contract implementation) may be needed. However, in many organization, they often choose to use only one of the three implementations to handle all their business needs. The following paragraphs explain each component of Fig. 1 in more detail.

1. Core Machine Learning: it is the implementation of machine learning algorithm in its native form. It often includes the following components, as described earlier in this paper: model initialization, model training, model validation, model scoring, model evaluation, model serialization, and model cleanup. Some steps are optional, as stated previously.

2. Server Layer Implementation: it is the implementation of the machine learning algorithm after code refactoring so that it can run on top of the Server Layer. In the context of this

paper, a Server Layer is a cloud-based computing environment that can:

I. Host one or more machine learning activities, including model training, scoring, and others.

II. Receive, organize, and disseminate data records to a specified machine learning task.

III. Handle scoring requests and other requests.

The Server Layer can have two running modes: Symmetric Multiprocessing (SMP) and Massively Parallel Processing (MPP). Therefore, the code refactoring often requires two steps accordingly: SMP code refactoring and MPP code refactoring.

2.1. SMP code refactoring includes the following components:

1) Thread utility functions (setting up the threads, running the threads, terminating the threads). See Pseudocode Snippet 1 in Appendix for an example of setting up the threads. Other thread utility functions can be written similarly.

2) Thread main functions (event handling, exception handling). See Pseudocode Snippet 2 in Appendix for an example of the thread main function.

3) Core Machine Learning Server Layer Implementation in SMP:

a) Model Initialization in SMP.

b) Model Training in SMP. See Pseudocode Snippet 3 in Appendix for an example of model training in SMP. Other Core Machine Learning Server Layer Implementations in SMP can be written similarly.

c) Model Validation in SMP.

d) Model Scoring in SMP.

e) Model Evaluation in SMP.

f) Model Serialization in SMP.

g) Model Clean-up in SMP.

2.2. MPP code refactoring includes the following components:

1) Cloud utility functions (setting up the cloud nodes, running the cloud nodes, terminating the cloud nodes).

2) Cloud main functions (MapReduce, event handling, exception handling).

3) Core Machine Learning Server Layer Implementation in MPP:

a) Model Initialization in MPP.

b) Model Training in MPP.

c) Model Validation in MPP.

d) Model Scoring in MPP.

e) Model Evaluation in MPP.

f) Model Serialization in MPP. See Pseudocode Snippet 4 in Appendix for an example of model serialization in MPP.

Other Core Machine Learning Server Layer Implementations in MPP can be written similarly.

g) Model Clean-up in MPP.

3. Streaming Layer Implementation: it is the implementation of the machine learning algorithm after code refactoring so that it can run on top of the Streaming Layer. In the context of this paper, a Streaming Layer is a computing environment that can:

I. Host one or more machine learning activities, including model training, scoring, and others.

II. Receive, organize and disseminate data events (one or multiple data records) to specified machine learning task in Sliding Windows; discard old data after use.

III. Organize and handle continuous queries, scoring requests, and other requests.

One of the main differences between the Streaming Layer and the Server Layer is that Streaming Layer handles data events on the fly (dynamically) and the Server Layer handles all the data at rest (statically). Another major difference is that Streaming Layer often discards old data to make room for new data while Server Layer usually does not discard old data.

The code refactoring for Streaming Layer Implementation includes the following components:

1) Sliding Window utility functions (setting up the sliding windows, running the sliding windows, terminating the sliding windows).

2) Sliding Window main function (data injecting, data discarding, event handling, exception handling).

3) Core Machine Learning Streaming Layer Implementation in Sliding Window:

a) Model Initialization in Sliding Window.

b) Model Training in Sliding Window.

c) Model Validation in Sliding Window. See Pseudocode Snippet 5 in Appendix for an example of model validation in Sliding Window. Other Core Machine Learning Streaming Layer Implementations in Sliding Window can be written similarly.

d) Model Scoring in Sliding Window.

e) Model Evaluation in Sliding Window.

f) Model Serialization in Sliding Window.

g) Model Clean-up in Sliding Window.

4. Smart Contract Implementation: it is the implementation of the machine learning algorithm after code refactoring so that it can run on top of the Smart Contract Layer. As mentioned previously, a Smart Contract is just an automated process. Once a machine learning algorithm is implemented as a Smart Contract and running on a blockchain, the automation problem can be solved. This is because when the predefined conditions are met, the automated process will be triggered and will run on the blockchain. In the context of

this paper, a Smart Contract Layer is a computing environment that can:

I. Host one or more machine learning activities in an automated and trustable way and cannot be canceled unilaterally.

II. Receive, organize, and disseminate data records to a specified machine learning task in an automated and trustable way and cannot be canceled unilaterally.

III. Handle scoring requests and other requests in an automated and trustable way and cannot be canceled unilaterally.

The code refactoring for Smart Contract Implementation includes the following components:

4.1. Smart Contract utility functions:

1) Reward [2] Smart Contract. It specifies all the details of the reward rules, including but not limited to: what are the winning criteria (best model wins), how much is the reward, how to distribute the reward, when will the reward be distributed, and what kind of digital wallet is required to receive the reward. The absence of this contract means there is no reward for winning models. See Pseudocode Snippet 6 in Appendix for an example of a Reward Smart Contract. Other smart contracts can be written similarly.

2) Model submission Smart Contract. It specifies the rules to allow participants to submit trained machine learning models. Models can be serialized and packed into either plain-text format or binary format. This contract shall disclose the details of the allowed model formats. Some commonly used plain-text formats include: Predictive Model Markup Language (PMML) [18], Portable Format for Analytics (PFA) [19], and Open Neural Network Exchange (ONNX) [20]. Binary format is designed and often optimized for computer run time and is not human-readable. Some participants might prefer to use certain binary formats to protect their intellectual property and to reduce scoring time. Therefore, the specification of a binary format, such as the ASTORE [21] format, is not an open-source standard, but a machine model based on a binary format can run on common platforms. See Pseudocode Snippet 7 in Appendix for an example of a Model Submission Smart Contract. Other smart contracts can be written similarly.

a) Fair-play Smart Contract. Most public and permissioned blockchains already have a mechanism in place to detect and prevent abusive behavior. However, a fair-play contract can still be used to enforce obligations and ensure a higher level of fairness. For example, this contract can require the reward to be deposited and secured so that the winning model will be rewarded. For another example, it can limit the number of submissions per day to avoid abusive and high-frequency submissions. See Pseudocode Snippet 8 in Appendix for an example of a Fair-play Smart Contract. Other smart contracts can be written similarly.

3) Smart Contract main function, see Pseudocode Snippet 9 in Appendix for an example of Smart Contract main function.

4) Core Machine Learning Smart Contract Layer Implementation:

a) Model initialization in Smart Contract. It initializes the system and ensures that only the participants who meet the system requirements can pass. For example, some machine learning algorithms require high-end CPU/GPU to find acceptable solutions. Machines without high-end CPU/GPU shall fail the system initialization contract.

b) Model training in Smart Contract. It prepares the training data set and makes it accessible to the public (for a public blockchain) or to permissioned participants (for a permissioned blockchain). It also specifies whether the training data set should be partitioned into a training portion and a validation portion. If partitioning should occur, it needs to specify how to partition the data set into these portions.

c) Model validation in Smart Contract. When a separate validation data set is available, it prepares the validation data set and makes it accessible to the public (for a public blockchain) or to permissioned participants (for a permissioned blockchain).

d) Model scoring in Smart Contract. It prepares the scoring data set and makes it accessible to the public (for a public blockchain) or to permissioned participants (for a permissioned blockchain). When there is no scoring data set, the training data set is used for scoring.

e) Model evaluation in Smart Contract. It specifies the rules for model evaluation and comparison so that submitted models can be ranked and a winning model can be selected.

f) Model serialization in Smart Contract.

g) Model clean-up in Smart Contract.

5. Server Layer APIs: are the APIs provided by the Server Layer. Currently, most Server Layer offerings come with SDK, which is a complete set of APIs that enable you to create applications to run on the Server Layer.

6. Streaming Layer APIs: are the APIs provided by the underneath Streaming Layer. Currently, most Streaming Layer offerings come with SDK, which is a complete set of APIs that enable you to create applications to run on the Streaming Layer.

7. Smart Contract APIs: are the APIs provided by the underlying Smart Contract Layer. Currently, most Smart Contract Layer offerings come with SDK, which is a complete set of APIs that enable you to create applications to run on the Smart Contract Layer.

8. Server Layer: In the context of this paper, a Server Layer is a cloud-based computing environment that can:

I. Host one or more machine learning tasks for model training and scoring.

II. Receive, save, organize, and disseminate data records to a specified machine learning task.

III. Organize and handle scoring requests.

9. Streaming Layer: In the context of this paper, a Streaming Layer is a computing environment that can:

  I. Host one or more machine learning tasks for model training and scoring.

  II. Receive, organize, and disseminate data events (one or multiple data records) to a specified machine learning task.

  III. Organize and handle continuous queries or scoring requests.

One of the main differences between the Streaming Layer and the Server Layer is that Streaming Layer handles data events on the fly (dynamically) and the Server Layer handles all the data at rest (statically). Another major difference is that the Streaming Layer often discards old data events to make room for new data, and the Server Layer does not discard old data.

10. Smart Contract Layer: In the context of this paper, a Smart Contract Layer is a computing environment that can:

  I. Host one or more machine learning tasks for model training and scoring in an automated and trustable way and cannot be canceled unilaterally.

  II. Receive, organize, and disseminate data records to a specified machine learning task in an automated and trustable way and cannot be canceled unilaterally.

  III. Organize and handle scoring requests in an automated and trustable way and cannot be canceled unilaterally.

11. Blockchain APIs: are the APIs provided by the underlying blockchain. Currently, many blockchain offerings come with SDK, which is a set of APIs that enable you to create applications to run on the blockchain. The Server Layer can obtain aggregated data at rest from the blockchain via blockchain APIs. The Streaming Layer can obtain instant data on the fly from the blockchain via blockchain APIs. The Smart Contract Layer can obtain data from the blockchain via blockchain APIs in a blockchain-native fashion.

12. Blockchain: is a continuously growing, single-linked list of immutable blocks (often secured using cryptography).

## IV. EXPERIMENT

In this section, a case study is presented to demonstrate how to use the proposed trustable machine learning to solve real problems. This case study uses Association Rule Mining (ARM, a subdomain of machine learning) to demonstrate how Trustable Machine Learning can be done with blockchain, and how to use it to do opioid prescription analysis to help combat the opioid crisis. The case study focuses on how to run ARM on top of the Server Layer to handle data at rest (coming from a blockchain all at one time).

### A. Data Set and Parameter Setup

The blockchain was created and managed by Kaleido [12], which is a software product from an early start-up company. Because of some technical issues, some parts of this study are incomplete. This permissioned blockchain is using protocol quorum/raft [17] as the consensus algorithm. There are three participating pharmacies in this blockchain: Pharmacy-A, Pharmacy-B and Pharmacy-C, as shown in Fig. 2. Each pharmacy logs its transactions to the blockchain. All data and transactions are synthetic. A transaction contains a patient ID and the patient's prescription of seven drugs. The drugs are all opioid drugs that we randomly grabbed from the Internet. A transaction is further divided into seven records. Each record contains a patient ID and one opioid drug. Some transactions are shown in Fig. 3. In total, there are 7,007 records of 1,001 patients and 20 different opioid drugs. The data were retrieved using SQL from the Kaleido blockchain to a Server Layer. The Server Layer used in this study is the SAS Cloud Analytical Server [21], which is a cloud-based computing Server Layer running on top of the SAS platform. ARM is applied with the following parameters: support $>= 20\%$, confidence $>= 70\%$, number of items in a rule $<= 3$.

Fig. 2. Blockchain setting.

Fig. 3. Synthetic data for the first two patients' prescriptions.

### B. Results and Remarks

Sixteen (16) rules are found and shown in Fig. 4. The first rule, "actiq & fentora ==> meperidine", means that if a patient is prescribed Actiq and Fentora, the patient is likely to be prescribed Meperidine later. ARM made this prediction on the basis of the frequency (20.17982010%, which is higher than the requested threshold of 20%) of this combination (Actiq, Fentora, and Meperidine) in the data set and its high confidence value (78.90625, which is higher than the requested threshold of 70%).

Fig. 4. Rules found by ARM in this synthetic case study.

This result does not make sense, because the entire data set is synthetic, and all the drugs are opioids. However, if given real prescription data, ARM should be able to find:

- Some "gateway" non-opioids that may lead to opioid prescriptions.
- Frequent combinations of opioid and non-opioids that are often prescribed together.
- Other interesting or surprising rules and analysis.

If the prescribing doctors' information or ID is available, more useful rules can be found. Because the data come from a blockchain and can be trusted, the findings can easily be verified and used for auditing, prosecution, or other purposes. Also, if ARM is implemented as a smart contract, this process can be automated. Policymakers and institutions can use this kind of information and details to gain more insights on and combat the opioid crisis. Combined with other techniques and results from other time periods, we can also do things like policy effect analysis which compares the current results with previous year and find out if the policy worked or not.

## V. CONCLUSION AND FUTURE RESEARCH

This paper presents a unified analytical framework for trustable machine learning and automation running with blockchain. The proposed framework can be used to solve the trustability and automation issues of machine learning algorithms. A preliminary experiment shows that the proposed framework is feasible. However, there are some problems associated with blockchain, and other problems that are often specific to the machine learning task. This section uses ARM as an example to discuss open problems and related research. In the context of this paper and ARM, three open problems are considered: privacy preservation, building incremental model and handling fast data stream. Privacy preservation is a big issue to many enterprise use cases, especially to those heavily regulated industries (such as healthcare, banking, insurance) that have data privacy laws. Some blockchains come with built-in privacy preservation solution therefore we don't have to worry about it. However, for some blockchains, privacy preservation is not guaranteed. The reason is that the top most design goals of blockchains are to provide data immutability and trustable replication. Those two goals are accomplished by replicating data throughout the network so that all participating nodes have the same copy which also ensures immutable data. As a result, privacy is not a built-in attribute in some blockchains. Blockchain is essentially an incremental model—there are some blocks sitting at rest and new blocks are added incrementally. Karp et al. [14] proposed a counting algorithm for finding frequent elements in data streams to build an incremental model. Their algorithm has an offline component and an online component, and four performance criteria were considered: amortized time, worst-case time, number of passes, and space. They also proved that the error is bounded. Chang and Lee [15] proposed an algorithm called estDec for association rule mining adaptively over an online data stream by decaying the effect of old transactions. The strength of this algorithm is that it can identify and highlight the recent change and trend. This makes sense in blockchain as well, because the old information may be no longer useful at present. Currently, the public blockchain (such as Bitcoin and Ethereum) data generation speed is not very fast. This means that the frequency counting [11] can be accurately done. However, blockchain data generation speed will soon increase, and it could exceed the threshold of accurate frequency counting. Therefore, we will need to add a fast data stream component to our software offering. Manku and Motwani [13] proposed sticky sampling and lossy counting algorithms for approximate frequency counts over fast data streams. Yu et al. [16] presented an algorithm for association rule mining from high speed transactional data streams. They claimed some advantages over the traditional false-positive mining approach. They argued that their proposed false-negative approach can effectively mine frequent itemsets with memory consumption bounded, while in the false-positive approach the number of false-positive frequent itemsets could increase exponentially, making the mining memory consumption potentially unbounded. This is a very interesting approach that is suitable for fast data stream situations in blockchain. In the future, all three open problems will be investigated.

REFERENCES

[1] S. Nakamoto, "Bitcoin: A peer-to-peer electronic cash system," retrieved online, May 2018.

[2] V. Buterin, "A next generation smart contract and decentralized application platform," retrieved online, May 2018.

[3] L. Lamport, R. Shostak, and M. Pease, "The Byzantine generals problem," ACM Trans. on Programming Languages and Systems, vol. 4, no. 3, pp. 382–401, 1982.

[4] I. Osipkov, E. Vasserman, N. Hopper, and Y. Kim, "Combating double-spending using cooperative P2P systems," doi:10.1109/ICDCS.2007.91, 2007.

[5] J. Hoepman, "Distributed double spending prevention," arXiv:0802.0832v1, 2008.

[6] A. Hodges and A. Turing, The Enigma. London: Burnett Books, 1983.

[7] R3 Corda, https://www.r3.com/, retrieved online, May 2018.

[8] Chain, https://chain.com/, retrieved online, May 2018.

[9] BigChainDB, https://www.bigchaindb.com/, retrieved online, May 2018.

[10] Hyperledger, https://www.hyperledger.org/, retrieved online, May 2018.

[11] R. Agrawal, T. Imieliński, and A. Swami, "Mining association rules between sets of items in large databases," Proceedings of the 1993 ACM SIGMOD International Conference on Management of Data, SIGMOD 1993.

[12] Kaleido, a ConsenSys company, https://kaleido.io/, retrieved online, May 2018.

[13] G. Manku and R. Motwani, "Approximate frequency counts over data streams," in Proceedings of the 2002 International Conference on Very Large Data Bases (VLDB '02), Hong Kong, pp. 346–357.

[14] R. Karp, C. Papadimitriou, and S. Shenker, "A simple algorithm for finding frequent elements in streams and bags," ACM Trans. Database Syst., vol. 28, pp. 51–55, 2003.

[15] J. Chang and W. Lee, "Finding recent frequent itemsets adaptively over online data streams," KDD '03, Washington, DC, pp. 487–492.

[16] J. Yu, Z. Chong, H. Lu, and A. Zhou, "False positive or false negative: Mining frequent itemsets from high speed transactional data streams," VLDB '04.

[17] https://en.wikipedia.org/wiki/Raft_(computer_science), retrieved 2018.

[18] R. Grossman, S. Bailey, A. Ramu, and X. Qin, "The management and mining of multiple predictive models using the predictive modeling markup language," Information and Software Technology, vol. 41, no. 9, pp. 589–595, 2002, DOI: 10.1016/S0950-5849(99)00022-1.

[19] J. Pivarski, C. Bennett, and R. Grossman, "Deploying analytics with the Portable Format for Analytics (PFA)," KDD '16.
[20] K. Hazelwood, et al., "Applied machine learning at Facebook: A datacenter infrastructure perspective," International Symposium on High-Performance Computer Architecture (HPCA), 2018.
[21] J. Wexler, S. Haller, and R. Myneni, "An overview of SAS Visual Data Mining and Machine Learning on SAS Viya," SAS Global Forum, 2017.
[22] A. Kurtulmus and K. Daniel, "Trustless machine learning contracts: Evaluating and exchanging machine learning models on the Ethereum blockchain," arXiv:1802.10185, 2018.
[23] H. Gilbert and H. Handschuh, "Security analysis of SHA-256 and Sisters," Selected Areas in Cryptography, 2003, pp. 175–193.

## APPENDIX: PSEUDOCODE

```
Function Setting_up_the_threads()
{
    status = OK;
    If (OK != Allocate_memory_for_threads() || OK != Allocate_memory_for_thread_events())
        status = OUT_OF_MEM and goto Finish;
    Init_thread_parameters();
Finish:
    If status =! OK
        Clean_up_the_threads();
    Return status;
}
```
Pseudocode Snippet 1: Setting_up_the_threads() Function Example

```
Function Thread_main()
{
    status = OK;
    push_exception_handler_to_stack();
    wait_for_thread_event
    {
        If (thread_event ==MODEL_INITIALIZATION)
            status = Model_initialization_at_thread_level(thread_event);
        Else if (thread_event ==MODEL_TRAINING)
            status = Model_training_at_thread_level(thread_event);
        Else if (thread_event ==MODEL_VALIDATION)
            status = Model_validation_at_thread_level(thread_event);
        Else if (thread_event ==MODEL_SCORING)
            status = Model_scoring_at_thread_level(thread_event);
        Else if (thread_event ==MODEL_EVALUATION)
            status = Model_evaluation_at_thread_level(thread_event);
        Else if (thread_event ==MODEL_SERIALIZATION)
        {
            status = Model_serialization_at_thread_level(thread_event);
            If (status == OK)
                status = Model_serialization_at_machine_level_for_all_threads();
        }
        Else if (thread_event ==MODEL_CLEAN_UP)
            status = Model_clean_up_at_thread_level(thread_event);
        Else
            Break;
    }
Finish:
    pop_exception_handler_from_stack();
    Return status;
}
```
Pseudocode Snippet 2: Thread_main () Function Example

```
Function Model_training_at_thread_level(thread_event)
{
    status = OK;
    If (OK != Allocate_memory_for_training_at_thread_level())
        status = OUT_OF_MEM and goto Finish;
    status = obtain_data_from_machine_memory();
    If (status == OK)
        status = train_the_model();
Finish:
    If status =! OK
        Clean_up_the_threads();
    Return status;
}
```
Pseudocode Snippet 3: Model Training in SMP Code Example

```
Function Model_serialization_at_cloud_level(node_instance)
```

```
{
    status = OK;

    If (node_instance == WORKER)
        status = Send_model_to_master_for_aggregation();
    Else if (node_instance == MASTER)
        status = Receive_model_from_worker_for_aggregation();

Finish:
    If (node_instance == MASTER || node_instance == WORKER)
        wait_until_model_aggregation_is_done();

    status = finish_writting_model_to_disk();
    return status;
}
```
Pseudocode Snippet 4: Model Serialization in MPP Code Example

```
Function Model_validation_in_sliding_window(data_stream)
{
    status = OK;

    If (data_stream == ERROR || data_stream == EMPTY)
        goto Finish;

    If (OK != Allocate_memory_for_validation_in_sliding_window())
        status = OUT_OF_MEM and goto Finish;

    While (obtain_validation_data_from_data_stream() == OK)
    {
        status = model_validation();
    }

Finish:
    If (data_stream == ERROR)
        status = ERROR;

    return status;
}
```
Pseudocode Snippet 5: Model Validation in Sliding Window Code Example

```
Function Reward_smart_contract()
{
    status = OK;

    If (is_my_digital_wallet_OK() == FALSE)
        status = BAD_WALLET and goto Finish;

    If (winning_inquiry() == TRUE && total_reward() > 0 && can_collect_reward() == TRUE )
        If (share_reward_with_others() == TRUE)
            status = collect_divided_reward();
        Else
            status = collect_all_reward();

Finish:
    return status;
}
```
Pseudocode Snippet 6: Reward Smart Contract Code Example

```
Function Model_submission_smart_contract(serialized_model)
{
    status = OK;

    If (serialized_model == EMPTY)
        status = NO_MODEL and goto Finish;

    If (plain_text_format_required() == TRUE)
    {
        If (pmml_format_required() == TRUE)
            status = submit_serialized_model_in_pmml();
        Else if (pfa_format_required() == TRUE)
            status = submit_serialized_model_in_pfa();
```

```
                    Else if (onnx_format_required() == TRUE)
                            status = submit_serialized_model_in_onnx();
                    Else
                            status = submit_serialized_model_in_other_plain_text_format();
            }
            Else
            {
                    If (astore_format_required() == TRUE)
                            status = submit_serialized_model_in_astore();
                    Else
                            status = submit_serialized_model_in_other_binary_format();
            }
Finish:
            return status;
}
```
Pseudocode Snippet 7: Model Submission Smart Contract Code Example

```
Function Fair_play_smart_contract()
{
            status = OK;

            If (shall_I_deposit_reward() == TRUE)
                    status = deposit_reward() and goto Finish;

            If (submission_limit_reached() == TRUE)
                    goto Finish;
            Else
                    status = try_to_train_and_submit_better_model();

Finish:
            return status;
}
```
Pseudocode Snippet 8: Fair-Play Smart Contract Code Example

```
Function Smart_contract_main()
{
            status = OK;
            push_exception_handler_to_stack();

            Fair_play_smart_contract();

            wait_for_event
            {
                    If (event ==MODEL_INITIALIZATION)
                            status = Model_initialization_in_smart_contract();
                    Else if (event ==MODEL_TRAINING)
                            status = Model_training_in_smart_contract();
                    Else if (event ==MODEL_VALIDATION)
                            status = Model_validation_in_smart_contract();
                    Else if (event ==MODEL_SCORING)
                            status = Model_scoring_in_smart_contract();
                    Else if (event ==MODEL_EVALUATION)
                            status = Model_evaluation_in_smart_contract();
                    Else if (event ==MODEL_SERIALIZATION)
                            status = Model_serialization_in_smart_contract();
                    Else if (event ==MODEL_CLEAN_UP)
                            status = Model_clean_up_in_smart_contract();
                    Else if (event == MODEL_SUBMISSION)
                            status = Model_submission_in_smart_contract();
                    Else
                            Break;
            }

            status = Reward_smart_contract();

Finish:
            pop_exception_handler_from_stack();
            Return status;
}
```
Pseudocode Snippet 9: Smart Contract Main Function Code Example